\documentclass{article}
\pdfoutput=1

\PassOptionsToPackage{numbers, compress}{natbib}
 \usepackage[final]{nips_2017}

\usepackage[utf8]{inputenc} 
\usepackage[T1]{fontenc}    
\usepackage{hyperref}       
\usepackage{url}            
\usepackage{booktabs}       
\usepackage{amsfonts}       
\usepackage{nicefrac}       
\usepackage{microtype}      

\usepackage{amsmath}

\usepackage{graphicx}
\usepackage{subcaption}
\newcommand\numberthis{\addtocounter{equation}{1}\tag{\theequation}}
\newcommand*\diff{\mathop{}\!\mathrm{d}}

\title{Bayesian LSTMs in medicine}

\author{
  Jos van der Westhuizen \hspace{3cm} Joan Lasenby \\
  University of Cambridge\\
  \texttt{\{jv365,jl221\}@cam.ac.uk} \\
}

\begin{document}

\maketitle

\begin{abstract}
The medical field stands to see significant benefits from the recent advances in deep learning. Knowing the uncertainty in the decision made by any machine learning algorithm is of utmost importance for medical practitioners. This study demonstrates the utility of using Bayesian LSTMs for classification of medical time series. Four medical time series datasets are used to show the accuracy improvement Bayesian LSTMs provide over standard LSTMs. Moreover, we show cherry-picked examples of confident and uncertain classifications of the medical time series. With simple modifications of the common practice for deep learning, significant improvements can be made for the medical practitioner and patient. 
\end{abstract}

\section{Introduction}
Life and death decisions are commonplace in the medical domain. When making medical decisions, doctors mostly evaluate multiple parameters and make decisions based on a complex mixture of intuition and assumptions. Machine learning has demonstrated groundbreaking performance in recent studies \citep{krizhevsky_imagenet_2012,mnih_human-level_2015,silver_mastering_2016,goodfellow_generative_2014} and shows promise as an augmentation to aid doctors in day-to-day care \citep{clifton_health_2015,lipton_learning_2015,zhang_when_2016}. One of the most promising current techniques is deep learning. For the specific class of temporal data that is ubiquitous in medicine, the branch of deep neural networks called Recurrent Neural Networks (RNNs), has yielded some of the best results \citep{lipton_learning_2015,choi_doctor_2015,jagannatha2016bidirectional,harutyunyan_multitask_2017}.

Although RNNs and other temporal models have shown much promise in analyzing sequential medical data, the models don't provide practitioners with a certainty measure of their decisions. Thus doctors have no quantitative measure of the importance they should place on the decisions made by their computational assistants. Clinicians typically determine the course of treatment given the current health status of the patient as well as some internal estimate of the outcome of possible future treatments. The effect of treatments for a given patient is non-deterministic (uncertain), and predicting the effect of a series of treatments over time compounds the uncertainty \citep{bennett_artificial_2013}. Uncertainty in medical decisions is of paramount importance.

Bayesian probability theory offers a mathematically grounded technique to reason about model uncertainty \citep{gal_dropout_2016}. However, these Bayesian techniques are often accompanied by a prohibitive computational cost. Previous research has explored the benefits of Bayesian techniques in medicine \citep{temko_eeg-based_2011,kononenko2001machine,meyfroidt_machine_2009,murphy_machine_2012,mani_medical_2014,ghassemi_multivariate_2015,guiza_grandas_gaussian_2006}.
However, these proposals do not harness the representative power exhibited by deep learning \citep{ongenae_time_2013}. Our work follows that of \citet{gal_theoretically_2015} to show that deep learning tools can be used as Bayesian models without changing the model for optimization.

But, do we not get confidence measures from the probabilities produced by the softmax function at the end of most neural networks? The probabilities obtained from Bayesian approaches is significantly different to the "probabilities" obtained from the softmax classifier \citep{kendall_bayesian_2015}. The softmax function provides estimates of the relative probabilities between classes, but not an overall measure of the model's uncertainty \citep{gal_dropout_2016}.

Our work demonstrates two key benefits of employing Bayesian deep learning: (i) an increase in the classification accuracy of medical signals, and (ii) a measure of confidence in the model decisions. Although conventional Bayesian approaches are computationally expensive, the implementation proposed here would enable online classification in a clinical setting.

\section{Related work} \label{sec:related}
\citet{lipton_learning_2015} made use of LSTMs to diagnose patients with 128 different codes (one code for each medical condition). Similarly, \citet{choi_doctor_2015} made use of gated recurrent units to predict medication and diagnosis codes. Both of these studies demonstrate the efficacy of LSTMs for sequential medical data, albeit on low-resolution (<0.0003~Hz) signals.

Bayesian Neural Networks (NNs) are a class of NNs which are able to model uncertainty \citep{denker1990transforming,mackay1992practical}. These models provide a variance (uncertainty) of the predictions by learning distributions over the weights. Often they are computationally expensive, increasing the number of model parameters without increasing model capacity significantly \citep{kendall_bayesian_2015}. Conventional Bayesian NNs mostly employ variational inference to approximate the posterior \citep{graves2011practical}.

\textit{Dropout} is a regularization technique commonly used in NNs to prevent overfitting and co-adaption of features \citep{srivastava_dropout:_2014}. The technique entails removing a percentage of random units within a network during each iteration of stochastic gradient descent. The standard approach is to rescale the weights at test time through multiplication of the learned weights by the probability of the weights being present during training, known as \textit{weight averaging}.

Rather conveniently, dropout can be used as approximate Bayesian inference over the weights of a network \citep{gal_bayesian_2015}, mitigating the computational complexity of Bayesian NNs. This is achieved by sampling from the network with random units removed at test time. Thus the NN does not require any additional parameters and a Bernoulli distribution is imposed over the weights. The samples can be considered as Monte Carlo samples obtained from the posterior distribution over models, giving rise to the name \textit{Monte Carlo (MC) dropout}. Using RNNs with MC dropout has seen success in \citep{gal_theoretically_2015} and in \citet{sennrich2016edinburgh}. 

Long Short-Term Memory (LSTM) RNNs are easier to train and perform better than standard RNNs \citep{hochreiter_long_1997}. Here we aim to demonstrate the efficacy of Bayesian LSTMs in medicine to improve accuracy and decrease the uncertainty in the final decisions that doctors make. \citet{fortunato_bayesian_2017} proposed a technique for obtaining uncertainty estimates using an adaptation of Bayes by Backprop \citep{graves2011practical}. Although the proposed technique yields accuracies superior to the technique in \citet{gal_theoretically_2015}, we choose to employ techniques proposed by the latter, which requires a smaller adaptation of commonly used techniques.

The Physionet/Computation in Cardiology 2016 Challenge provides an appropriate dataset for benchmarking the performance of LSTMs \citep{liu2016open,clifford2016classification}. This comprehensive dataset was recently collected, is multi-center, and has multiple reported performance scores. The dataset comprises 4,430 heart sound recordings lasting from several seconds to over 100s with a resolution of 2~kHz. The data have long and short-term features paramount for classification of the signal. Moreover as detailed in \citet{springer2016automated} accurate classification of these signals is vital in developing communities. Among the top performing techniques for the official challenge were convolutional NNs, an ensemble of support vector machines, regularized NNs, and random forests. \citet{harutyunyan_multitask_2017} proposed an easy to use benchmark system for medical data that is based on the Medical Information Mart for Intensive Care (MIMIC-III). The benchmark includes four different medical tasks based on low-resolution data. However, owing to more information being available in medical signals collected at higher resolution we feel it is important to also benchmark temporal models on the latter.

\section{Methods}
The LSTM implemented is based on the model described in \citet{hochreiter_long_1997} and implemented in Tensorflow \citep{tensorflow2015-whitepaper}. Each cell in the LSTM has \textit{input}, \textit{output}, \textit{forget}, and \textit{input modulation} gates $ \textbf{i}, \textbf{o}, \textbf{f}$ and $\textbf{g} $.
\begin{align*}
&\textbf{i}=\sigma(\textbf{h}_{t-1}\textbf{U}_i+\textbf{x}_t\textbf{W}_i +\textbf{b}_i)         &  &\textbf{f}=\sigma(\textbf{h}_{t-1}\textbf{U}_f+\textbf{x}_t\textbf{W}_f +\textbf{b}_f)            \\
&\textbf{o}=\sigma(\textbf{h}_{t-1}\textbf{U}_o+\textbf{x}_t\textbf{W}_o +\textbf{b}_o)         &  &\textbf{g}=\tanh(\textbf{h}_{t-1}\textbf{U}_g+\textbf{x}_t\textbf{W}_g +\textbf{b}_g) \\
&\textbf{c}_t=\textbf{f}\odot\textbf{c}_{t-1}+\textbf{i}\odot\textbf{g}   &  
&\textbf{h}_t=\textbf{o}\odot\tanh(\textbf{c}_t)   \numberthis \label{lstm_eq}   
\end{align*}
The internal state $ \textbf{c}_t $ is referred to as \textit{cell} and is updated additively. The non-linear sigmoid activation is represented by $ \sigma $, and $\textbf{W}_* $ and $ \textbf{U}_* $ are the input and hidden weight matrices respectively with biases $ \textbf{b}_* $. We re-parameterize the model to have a single weight matrix $ \textbf{W}_l $ for layer $ l $. For a specific layer, the input to each gate's non-linearity is then computed by the single matrix multiplication:
\begin{equation*}
\begin{bmatrix}
\textbf{x}_t & \textbf{h}_{t-1} & 1
\end{bmatrix}\textbf{W}_l
=
\begin{bmatrix}
\textbf{x}_t & \textbf{h}_{t-1} & 1
\end{bmatrix}
\begin{bmatrix}
\textbf{W}_i & \textbf{W}_f & \textbf{W}_o & \textbf{W}_g \\
\textbf{U}_i & \textbf{U}_f & \textbf{U}_o & \textbf{U}_g \\ 
\textbf{b}_i & \textbf{b}_f & \textbf{b}_o & \textbf{b}_g
\end{bmatrix}
\end{equation*}
\small
\begin{align*}
=
\begin{bmatrix}
\textbf{x}_t\textbf{W}_i+\textbf{h}_{t-1}\textbf{U}_i+\textbf{b}_i & 
\textbf{x}_t\textbf{W}_f+\textbf{h}_{t-1}\textbf{U}_f+\textbf{b}_f &
\textbf{x}_t\textbf{W}_o+\textbf{h}_{t-1}\textbf{U}_o+\textbf{b}_o & 
\textbf{x}_t\textbf{W}_g+\textbf{h}_{t-1}\textbf{U}_g+\textbf{b}_g 
\end{bmatrix}
\numberthis \label{eq:reparam}
\end{align*}
\normalsize
with the resulting vector partitioned into the sum terms for input to the non-linearities in Equation \ref{lstm_eq}. This results in a single distribution being placed over one weight matrix $ \textbf{W}_l $ when applying dropout. The implication of the single weight matrix is a faster forward-pass with slightly diminished results \citep{gal2016uncertainty}.

\subsection{Bayesian LSTM} \label{sec:bayes}
We perform approximate inference in a Bayesian LSTM \citep{gal_dropout_2016} by using dropout \citep{srivastava_dropout:_2014}. Therefore, dropout can be considered as a way of getting samples from the posterior distribution of models. This technique is linked to variational inference in a Bayesian NN with Bernoulli distributions over the network's weights \citep{gal_dropout_2016}. We leverage this method to perform Bayesian inference with LSTMs.

We are interested in finding the posterior distribution of the LSTM weights, $ \omega $, given the observed labels $ \textbf{Y} $, and data $ \textbf{X} $.
\begin{equation}
p(\omega|\textbf{X},\textbf{Y})
\end{equation}
This posterior distribution is not tractable in general, and we use variational inference to approximate it \citep{kendall_bayesian_2015,gal_dropout_2016,denker1990transforming,graves2011practical}. This allows us to learn over the network's weights, $ \omega~=~\{\textbf{W}_1,...,\textbf{W}_L\}$, by minimizing the reverse Kullback Leibler (KL) divergence between this approximating distribution and the full posterior;
\begin{equation}
KL(q(\omega)||p(\omega|\textbf{X},\textbf{Y}))
\end{equation}
where $ q(\omega) $ is a distribution over matrices whose columns are randomly set to zero. For the LSTM, these matrices, $ \textbf{W}_l $ (Equation \ref{eq:reparam}), are all the weights on a single layer $ l $ and each matrix $ \textbf{W}_l $ has dimensions $ K_{l-1} \times K_l $. $ q(\omega) $ can be defined as:
\begin{align*}
&\textbf{W}_l =\textbf{M}_l \cdot \mathrm{diag}([z_{l,k}]^{K_l}_{k=1}) 
\\
&z_{l,k} \sim \mathrm{Bernoulli}(p_l) \quad \mathrm{for} \quad l = 1,...,L, \quad k = 1,...,K_{l}
\numberthis \label{eq:dropout}
\end{align*}
given some probabilities $ p_l $ and matrices $ \textbf{M}_l $ as variational parameters. The binary variable $ z_{l,k} = 0$ corresponds to the output of a unit $ k $ in layer $ l $ being dropped. Note that we can left multiply the matrices $ \textbf{M}_l $ with a similar diagonal matrix in Equation \ref{eq:dropout} to apply dropout over the rows (unit inputs).

Given the LSTM definitions in \ref{lstm_eq}, we can re-write the operation (omitting biases for brevity) as a function $ f_h $:
\begin{align*}
\textbf{h}_t &= f_h(\textbf{x}_t,\textbf{c}_{t-1},\textbf{h}_{t-1})
\\ &=
\sigma(\textbf{h}_{t-1}\textbf{U}_o+\textbf{x}_t\textbf{W}_o)\odot\tanh(\sigma(\textbf{h}_{t-1}\textbf{U}_f+\textbf{x}_t\textbf{W}_f)\odot\textbf{c}_{t-1}
\\ & \qquad +\sigma(\textbf{h}_{t-1}\textbf{U}_i+\textbf{x}_t\textbf{W}_i)\odot\sigma(\textbf{h}_{t-1}\textbf{U}_g+\textbf{x}_t\textbf{W}_g)) \numberthis \label{eq:h}
\end{align*}
where $ \textbf{c}_{t-1} $ is the hidden unit memory from the previous time step and is determined by a recursive function on $ \textbf{h}_{t-2} $.
The output can be defined as $ f_y(\textbf{h}_T)=\textbf{h}_T\textbf{W}_y+\textbf{b}_y$. This LSTM can be viewed as probabilistic model by regarding the weights, $ \omega = \{\textbf{W}_*,\textbf{U}_*,\textbf{b}_*\}$ to be random variables (following normal prior distributions). The functions are written as $ f_h^\omega $ and $ f_y^\omega $ to emphasize the dependence on $ \omega $. Approximating the posterior distribution $ q(\omega) $ we have:
\begin{align*}
\int q(\omega)\log p(\textbf{y}|f_y^\omega(\textbf{h}_T))\diff \omega
&=
\int q(\omega)\log p\left(\textbf{y}|f_y^\omega(f_h^\omega(\textbf{x}_T,\textbf{c}_{T-1},\textbf{h}_{T-1}))\right)\diff \omega
\\ &=
\int q(\omega)\log p\big(\textbf{y}|f_y^\omega(f_h^\omega(\textbf{x}_T,\textbf{c}_{T-1},f_h^\omega(...f_h^\omega(\textbf{x}_1,\textbf{c}_0,\textbf{h}_0)...)))\big)\diff \omega
\end{align*}
with $ \textbf{h}_0 = \textbf{c}_0 = 0 $. We approximate this via MC integration with a single sample:
\begin{equation*}
\approx \log p\big(\textbf{y}|f_y^{\hat{\omega}}(f_h^{\hat{\omega}}(\textbf{x}_{T},\textbf{c}_{T-1},f_h^{\hat{\omega}}(...f_h^{\hat{\omega}}(\textbf{x}_1,\textbf{c}_0,\textbf{h}_0)...)))\big), \qquad \qquad \hat{\omega}\sim q(\omega) 
\end{equation*}
resulting in an unbiased estimator of each sum term. Our minimization objective then becomes:
\begin{equation} \label{eq:objective}
\mathcal{L} \approx - \sum_{i=1}^{N}\log p\big(\textbf{y}_i|f_y^{\hat{\omega}_i}(f_h^{\hat{\omega}_i}(\textbf{x}_{i,T},\textbf{c}_{T-1},f_h^{\hat{\omega}_i}(...f_h^{\hat{\omega}_i}(\textbf{x}_{i,1},\textbf{c}_0,\textbf{h}_0)...)))\big) + \mathrm{KL}(q(\omega)||p(\omega))
\end{equation}
From Equation \ref{eq:dropout}, we define our approximating distribution $ q(\omega) $ to factorize over the weight matrices and their columns in $ \omega $ \citep{gal_theoretically_2015}. For each layer $ l $ every weight matrix column $ \textbf{w}_{lk} $ the approximating distribution is:
\begin{equation}
q(\textbf{w}_{lk})=p\mathcal{N}(\textbf{w}_{lk};\textbf{m}_{lk},\sigma^2I)+(1-p)\mathcal{N}(\textbf{w}_{lk};\textbf{0},\sigma^2I)
\end{equation}
with $ \textbf{m}_{lk} $ variational parameter (column vector), small $ \sigma^2 $, and $ 1-p $ the dropout probability provided in advance.  We optimize over the variational parameters of the random weight matrices; these correspond to the LSTM weight matrices in the standard view. The KL term in Equation \ref{eq:objective} can be approximated as $ \sum_{l=1}^{L}\frac{p_l}{2N}(||\textbf{M}_l||_2^2 +||\textbf{b}_l||_2^2)$, summing over the variational parameters $ \textbf{M}_l $ of each weight matrix $ \textbf{W}_l $ in our model (each composed of weight vectors $ \textbf{m}_{lk} $) \citep{gal_dropout_2016}.

Evaluating the model output $ f_y^{\hat{\omega}}(\cdot) $ with sample $ \hat{\omega} $ corresponds to randomly zeroing (masking) columns in each weight matrix $\textbf{W}_l$ during the forward pass -- i.e. performing dropout. Further, our objective $ L $ is identical to that of the standard LSTM. In the LSTM setting with a sequence input, each weight matrix row is randomly masked.

Predictions can be approximated using the standard forward pass for LSTMs, i.e., propagating the mean of each layer to the next (\textit{standard dropout approximations}), or by approximating the posterior with $ q(\omega) $ for a new input $ \textbf{x}^* $,
\begin{align*}
p(\textbf{y}^*|\textbf{x}^*,\textbf{X}, \textbf{Y}) &\approx \int p(\textbf{y}^*|\textbf{x}^*,\omega)q(\omega)\diff \omega \\
&\approx \frac{1}{K} \sum_{k=1}^{K}p(\textbf{y}^*|\textbf{x}^*,\hat{\omega}_k)
\end{align*}
with $ \hat{\omega}_k \sim q(\omega) $, i.e. by performing dropout at test time and averaging the results (\textit{MC dropout}).

\citet{gal_theoretically_2015} emphasizes that for each sample $ \textbf{x}_i $ a single realization $ \hat{\omega}_i = \{\hat{\textbf{W}}_*^i,\hat{\textbf{U}_*^i},\hat{\textbf{b}}_*^i\}$ is sampled, and that element in the sequence $ \textbf{x}_i =[\textbf{x}_{i,1},...,\textbf{x}_{i,T}]$ is passed through a function with the same parameters $ f_h^{\hat{\omega}_i} $. This is referred to as \textit{Variational} dropout. Intuitively, having the same dropout mask per sequence element makes sense from a recurrent and Monte Carlo integration approximation perspective. However, empirically we found that \textit{naive} dropout, with different samples at each time step $ {\hat{\omega}_{i,1},...,\hat{\omega}_{i,T}} \sim q(\omega) $ still improves the classification performance when using MC dropout compared to the standard dropout approximation.

When sampling a different $ \hat{\omega}_{i,t} $ for each recursion function $ f_h^{\hat{\omega}_{i,t}} $ (i.e. each time step in $ \textbf{x}_i $) in Equation \ref{eq:objective}, the function is no longer strictly recursive. At each level of recursion a different function $ f_h^{\hat{\omega}_{i,t}} $ is applied to the $ t^{\text{th}} $ element of $ \textbf{x}_i $. However, if an optimum is reached during training, each sample $ \hat{\omega}_{i,t} $ would produce a similar function $ f_h^{\hat{\omega}} $, making Equation \ref{eq:objective} approximately recursive. With naive dropout the minimization objective becomes
\begin{align*}
\mathcal{L} \approx &- \sum_{i=1}^{N}\log p\big(\textbf{y}_i|f_y^{\hat{\omega}_{i,k}}(f_h^{\hat{\omega}_{i,T}}(\textbf{x}_{i,T},\textbf{c}_{T-1},f_h^{\hat{\omega}_{i,T-1}}(...f_h^{\hat{\omega}_{i,1}}(\textbf{x}_{i,1},\textbf{c}_0,\textbf{h}_0)...)))\big) \\ &+ \mathrm{KL}(q(\omega)||p(\omega)) \numberthis \label{eq:naive}
\end{align*}
where $ \hat{\omega}_{i,k} $ represents an arbitrary dropout mask for the linear mapping $ f_y $ defined earlier. $ T $ represents the number of elements in $ \textbf{x}_i $. The first term in Equation \ref{eq:naive} pushes the posterior $ q(\omega) $ towards a Dirac delta function in order to have the function be the same at each time step.

\begin{figure}[!htbp]	
	\begin{subfigure}{.49\textwidth}
		\centering
		\includegraphics[width=\linewidth]{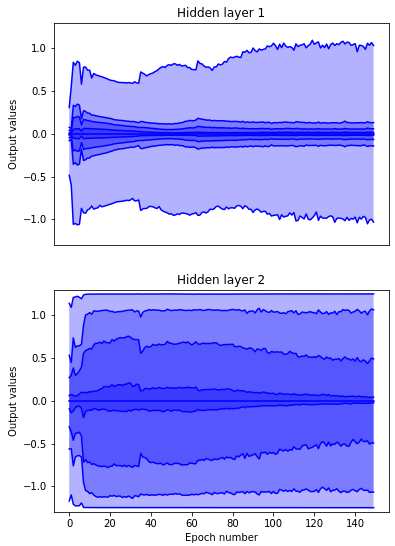}
		\caption{Naive dropout}
	\end{subfigure}
	\hspace{0.1cm}
	\begin{subfigure}{.49\linewidth}
		\centering
		\includegraphics[width=\linewidth]{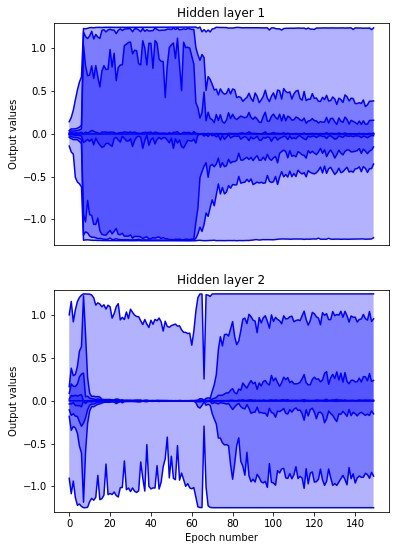}
		\caption{Variational dropout}
	\end{subfigure}
	\caption{Hidden unit output distributions for the naive and variational dropout approaches. From top to bottom, the lines represent the maximum, $ 93^{rd} $, $ 86^{th} $, $ 69^{th} $, $ 31^{st} $, $ 14^{th} $, $ 7^{th} $ percentiles, and the minimum. The output values exceed the (-1,1) range due to the Tensorflow implementation of dropout scaling the weights by $ 1/(keep\ probability) $ during training \citep{tensorflow2015-whitepaper}.}
	\label{fig:samples}
\end{figure}

The difference between the variational and naive dropout approaches is depicted in Figure~\ref{fig:samples}. The distributions of the hidden outputs (Equation \ref{eq:h}) after dropout (sampled parameters) are plotted over 150 epochs for a model trained on the MNIST dataset described in Section \ref{sec:exp}. The graphs show the percentiles of the hidden layer outputs over all time steps for the same arbitrary input sample at each epoch. Although both approaches result in similar performance (Table \ref{tab:acc}), the converged hidden output distributions are quite different. In accordance with the hypothesis above, the naive approach results in a narrow distribution on the first layer with a standard deviation of 0.1224 compared to the variational approach (0.2818). The second layers in both approaches seem to counter the distributions of the first layers -- the wide range of parameter exploration in the first layer of the variational approach has a concurrent narrow band of exploration in the second layer. During experimentation, it was found that the distribution of the variational approach is the same for any training simulation, where the distributions over time for the naive approach would vary between different training simulations.

Intuitively the variational dropout should be easier to train than the naive approach because the naive approach is not strictly recursive during the initial stages of training. The inherent leakiness of the LSTM memory \citep{neil_phased_2016} could be one reason why the LSTMs converge during training with naive dropout. The leakiness of the network results in bad samples from the posterior to be leaked (forgotten over time).

\subsection{Experimental implementation} \label{sec:exp}
We demonstrate the efficacy of Bayesian LSTMs by means of 5 datasets described in the following sections. The same LSTM model with a different architecture was used for each dataset (see the following sections for details). The outputs of the last hidden layer were linearly mapped to the output dimension. The resulting vectors were then average pooled before being subjected to the softmax function. A validation set was used in each case for early stopping of training. Dropout was used on only the input and output LSTM connections. Optimization was performed with Adam \citep{kingma_adam:_2014}, a learning rate of 0.01, and a minibatch size of 256. The standard and Bayesian LSTMs referred to hereafter are the same models, but for the Bayesian LSTM, MC dropout was used during testing to provide a measure of uncertainty.

\subsubsection{MNIST}
The MNIST handwritten digit dataset \citep{lecun1998mnist} provided by Tensorflow \citep{tensorflow2015-whitepaper} was processed in scanline order \citep{cooijmans2016recurrent}. The model architecture was 2 hidden layers with 128 units in each. A dropout value ($ 1-keep\ probability $) of 0.2 was used.

\subsubsection{MIT-BIH arrhythmia dataset}
This dataset contains 48 half-hour excerpts of electrocardiogram (ECG) recordings from 47 patients \citep{moody2001impact,goldberger_physiobank_2000}. The 5 heartbeat classes selected from the database were: normal beat, right bundle branch block beat, left bundle branch block beat, paced beat, and premature ventricular fibrillation. Single heart beats were extracted using the Pan-Tompkins algorithm \citep{pan_real-time_1985}, which has a reported accuracy of 0.99 on this dataset. The resulting dataset contained 106,848 samples of 216 time steps at 360~Hz. A random split of 50:40:10 (train:test:validation) was used. A model with a single hidden layer of 128 units and a dropout probability of 0.3 was used.

\subsubsection{Physionet/Compute in cardiology challenge 2016}
Of the 4,430 phonocardiogram (PCG) recordings in this dataset (see Section \ref{sec:related}), 3,126 were provided for training. The 301 validation samples (selected by the challenge organizers) were extracted from the training dataset. Each PCG signal was normalized independently to have a zero mean and unit standard deviation. Thereafter each signal was decimated to a frequency of 1~kHz. Owing to LSTMs not being able to handle long sequences \citep{neil_phased_2016}, we segmented the signals into samples with a length of at most 1000 time steps.

The data were provided with 2 classes; normal and abnormal heart beats. During online evaluation for the challenge, the models are allowed to classify a signal into a third class; noisy, resulting in a lower penalty on the model's score compared to an incorrect classification. To determine the class of a signal we first averaged the softmax probabilities over all the segments of the signal. For the standard LSTM we then classified a signal as noisy if the averaged softmax probabilities were between 0.45 and 0.55. For the Bayesian LSTM the signal was classified as noisy if the standard deviation (averaged over all the signal's segments) was higher than 0.13.

The online submission imposed a strong computational constraint on the model, with the virtual machine for the scoring having a single CPU core and 2GB of RAM. A model with 2 hidden layers of 128 units and a dropout probability of 0.25 was used. Model performance was evaluated by means of online submission that returns a score based on the specificity and sensitivity \citep{clifford2016classification}.

\subsubsection{Neonatal intensive care unit dataset}
This dataset contains the first 48 hours of vital signs for 3 neonatal intensive care unit (NICU) patients collected as part of the study by \citet{sortica_da_costa_complexity_2017}. The signals used for analysis were ECG, blood pressure, and oxygen saturation. The data were segmented into samples with a length of 200 time steps at 60~Hz, resulting in a total of 134,812 samples from 3 different classes: normal, dying, and intraventricular hemorrhage. Oxygen saturation values are the second-long averages, and clinicians were consulted to establish factors that scale the inputs to range from approximately 0 to 1. The employed model had a single hidden layer of 64 units and a dropout probability of 0.1. A random split of 50:40:10 was used.

\subsubsection{Traumatic brain injury dataset} \label{sec:tbi}
Data were collected from traumatic brain injury (TBI) patients as part of a larger study directed by the Department of Clinical Neurosciences at Addenbrookes. The dataset contains 19 variables recorded for 101 patients of which 34 were females, and the age ranged from 15 to 76. The dynamic variables comprised 5s averaged values for intracranial pressure (ICP), cerebral perfusion pressure, arterial blood pressure, heart rate, respiratory rate, systolic and diastolic blood pressure; the 5s amplitudes of arterial blood pressure, respiratory rate, and respiratory pulse; the minimum and maximum of ICP over the 5s; the peak-to-peak timing values for arterial blood pressure and ICP; the slow wave ICP; and the pressure-reactivity index values \citep{czosnyka1997continuous}. The static variables include age and gender. The duration of the recorded signals ranged from 1h to 12 days. The patients were classified according to the Glasgow Outcome Scale (GOS) \citep{jennett_assessment_1975}, providing a number between 1 and 5 to patients based on their health status 6 months after admission to the intensive care unit, with 5 being a good outcome and 1 being death. This dataset only contained patients with a GOS of 1 or 5. A random split of 50:40:10 was used. The model had a single hidden layer of 128 units and a dropout probability of 0.4. This dataset has a lower resolution than those introduced earlier and is used to demonstrate that the Bayesian approach is also beneficial for lower resolution longitudinal data.

\section{Results}
Table \ref{tab:acc} summarizes the results for the datasets analyzed in this study. The values shown are the averages for 10 runs. For the Bayesian LSTM 100 samples were used for MC dropout. Using MC dropout at test time improved the model accuracy on all the datasets, even though naive dropout was employed. In brackets we show the accuracies yielded for the variational dropout approach on the MNIST and MIT-BIH dataset. The variational approach significantly improved the accuracies for the MIT-BIH dataset, but yielded lower accuracies for the MNIST dataset. For the best model on the Physionet dataset the sensitivity and specificity values obtained were 0.675 and 0.880 for the standard LSTM, and 0.707 and 0.889 for the Bayesian LSTM respectively.

\begin{table}[!htbp]
	\centering
	\caption{Model Accuracies}
	\label{tab:acc}
	\begin{minipage}{.6\textwidth}
		\centering
		\begin{tabular}{p{3cm}p{2cm}p{2cm}}
			\toprule
			Dataset	& Standard LSTM & Bayesian LSTM  \\ \midrule
			MNIST	& 0.9889 (0.987) & \textbf{0.9891} (0.9879)\\
			Physionet 2016	\footnote{Online score, not accuracy.\\ \indent Values in brackets are the accuracies using variational dropout.}	& 0.778& \textbf{0.798}\\ 
			MIT-BIH		& 0.9815 (0.98463)& \textbf{0.9834} (0.98468)\\ 
			NICU		& 0.9972& \textbf{0.9979}\\ 
			TBI		& 0.9449& \textbf{0.9521}\\ 
			\bottomrule
		\end{tabular}
		\vspace{-0.75\skip\footins}
		\renewcommand{\footnoterule}{}
	\end{minipage}
\end{table}

\begin{figure}[!htbp]	
	\begin{subfigure}{.42\textwidth}
		\centering
		\includegraphics[width=\linewidth]{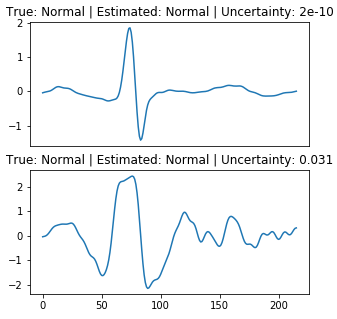}
		\caption{MIT-BIH ECG}
	\end{subfigure}%
	\begin{subfigure}{.42\linewidth}
		\centering
		\includegraphics[width=\linewidth]{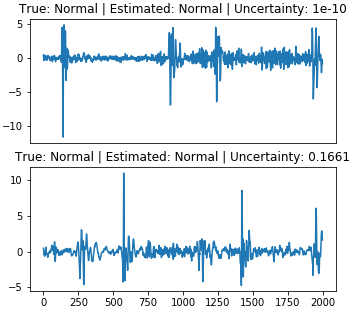}
		\caption{Physionet PCG}
	\end{subfigure}
	\begin{subfigure}{.14\linewidth}
		\centering
		\includegraphics[width=\linewidth]{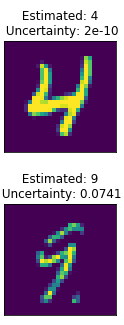}
		\captionsetup{skip=8pt}
		\caption{MNIST}
	\end{subfigure}
	\begin{subfigure}{.49\linewidth}
		\centering
		\includegraphics[width=\linewidth]{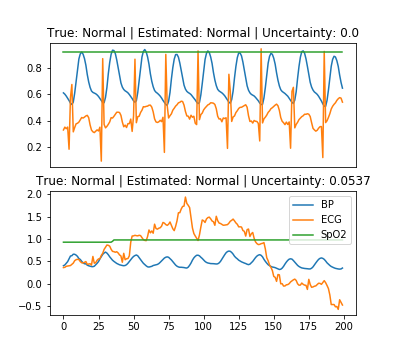}
		\caption{NICU}
	\end{subfigure}
	\begin{subfigure}{.49\linewidth}
		\centering
		\includegraphics[width=\linewidth]{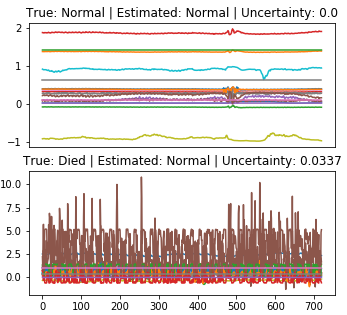}
		\caption{TBI}
	\end{subfigure}
	\centering
	\caption{Examples of confident classifications (top row) and uncertain classifications (bottom row) by the Bayesian LSTM on the different datasets. The medical samples displayed have been normalized and segmented. The NICU samples comprise the ECG, blood pressure (BP), and oxygen saturation signals (SpO2). Refer to Section \ref{sec:tbi} for details about the TBI signals.}
	\label{fig:uncertainties}
\end{figure}

As mentioned before, using a Bayesian LSTM for the classification of medical time series provides the imperative benefit of a confidence measure alongside the estimated class. In Figure \ref{fig:uncertainties} we juxtapose confident and uncertain Bayesian LSTM classified medical signals from the datasets analyzed in this study. It should be noted that for standard LSTMs, only the estimated class is produced as output. The figure shows that the model is uncertain when the signals look abnormal or noisy. The uncertainty value indicates when practitioners should further investigate signals and could help researchers understand how LSTM models work.

\section{Discussion}
The model yielded performance slightly below the benchmark performance on the Physionet 2016 challenge dataset \citep{clifford2016classification}. We believe that the LSTMs have the capacity to compete with the benchmark models for the Physionet 2016 Challenge. However, LSTMs are known to have poor performance on signals longer than 1000 time steps \citep{neil_phased_2016}. The original signals had to be split into subsegments of 1000 steps each. When splitting medical signals such as these, the subsegments of the original signal could be indicative of a different class, and assigning them with the same label as the original will confuse the model during training. Moreover, the strong computational constraints (single CPU core, and processing time limit) imposed by the competition does not allow for an LSTM model that has sufficient explanatory power. The model performance on the MNIST dataset is similar to that found in \citet{cooijmans2016recurrent} and \citet{zhang2016architectural}, 0.989 and 0.981 respectively.

As a practical guide to the implementation of dropout, our study found that ideal keep probabilities should be larger than 0.8 (dropout < 0.2) for LSTMs. LSTMs were found to converge to poor optimums and even overfit strongly when using keep probabilities around 0.5. Although higher keep probabilities result in weaker Bayesian uncertainties for the proposed implementations, the yielded variance still provides sufficient measures of confidence. Moreover, MC dropout is more computationally expensive than standard weight averaging, but owing to the samples being independent, it is a highly parallelizable method.

\section{Conclusion}
This study showed that a simple adaptation of the conventional deep learning technique for time series can (i) provide a vital additional output for quantifying model decisions, and (ii) improve model accuracy. Furthermore, we showed examples of applying this simple LSTM adaptation to medical data, where the contribution from a model confidence measure is greatly beneficial. In this work, we only focused on epistemic uncertainty - model uncertainty which can be explained away given enough data \citep{kendall_what_2017}. Methods for quantifying aleatoric uncertainty -- uncertainty inherent in observations could also provide valuable benefits to the medical machine learning field.

\bibliographystyle{apalike}
\bibliography{References/references,References/My_Library} 

\end{document}